\documentclass{article}
\usepackage{ijcai25}

\usepackage{times}
\usepackage{soul}
\usepackage{url}
\usepackage[hidelinks]{hyperref}
\usepackage[utf8]{inputenc}
\usepackage[small]{caption}
\usepackage{graphicx}
\usepackage{amsmath}
\usepackage{amsthm}
\usepackage{booktabs}
\usepackage[switch]{lineno}
\usepackage{amssymb}
\usepackage{color}

\usepackage[utf8]{inputenc} 
\usepackage[T1]{fontenc}    
\usepackage{hyperref}       
\usepackage{url}            
\usepackage{booktabs}       
\usepackage{amsfonts}       
\usepackage{nicefrac}       
\usepackage{microtype}      
\usepackage{xcolor}         
\usepackage{times}
\usepackage{epsfig}
\usepackage{graphicx}
\usepackage{amsmath}
\usepackage{amssymb}
\usepackage{algorithmic}
\usepackage[linesnumbered,ruled,vlined]{algorithm2e}
\usepackage{multirow}
\usepackage{booktabs}
\usepackage{siunitx}
\usepackage{bm}
\usepackage{amsfonts}
\usepackage{color}

\usepackage{dsfont}

\title{Noise Optimized Conditional Diffusion for Domain Adaptation}

\author{
    Lingkun Luo\textsuperscript{1},
    Shiqiang Hu\textsuperscript{1},
    Liming Chen\textsuperscript{2,3}
    \affiliations
    \textsuperscript{1}School of Aeronautics and Astronautics, Shanghai Jiao Tong University, Shanghai, China \\
    \textsuperscript{2}LIRIS, CNRS UMR 5205, Ecole Centrale de Lyon, France \\
    \textsuperscript{3}Institut Universitaire de France (IUF), France \\
    \emails
    lolinkun@gmail.com, sqhu@sjtu.edu.cn, liming.chen@ec-lyon.fr
\thanks{\textcolor{blue}{This work has been accepted by the International Joint Conference on Artificial Intelligence (IJCAI 2025).} This work was supported by the National Natural Science Foundation of China(62006152),  Natural Science Foundation of Shanghai (24ZR1434400), and China Aviation Science Foundation (2022Z071057002)．  Prof.Liming Chen in this work was in part supported by the French Research Agency, l’Agence Nationale de Recherche (ANR), through the projects Chiron (ANR-20-IADJ-0001-01), Aristotle (ANR-21-FAI1-0009-01), and Astérix (ANR-23-EDIA-0002) as well as the French national investment prioritary program PSPC FAIR WASTE project.}
}

\DeclareUnicodeCharacter{FF0E}{.}

\begin{document}
\maketitle
\begin{abstract}
Pseudo-labeling is a cornerstone of Unsupervised Domain Adaptation (UDA), yet the scarcity of High-Confidence Pseudo-Labeled Target Domain Samples (\textbf{hcpl-tds}) often leads to inaccurate cross-domain statistical alignment, causing DA failures. To address this challenge, we propose \textbf{N}oise \textbf{O}ptimized \textbf{C}onditional \textbf{D}iffusion for \textbf{D}omain \textbf{A}daptation (\textbf{NOCDDA}), which seamlessly integrates the generative capabilities of conditional diffusion models with the decision-making requirements of DA to achieve task-coupled optimization for efficient adaptation. For robust cross-domain consistency, we modify the DA classifier to align with the conditional diffusion classifier within a unified optimization framework, enabling forward training on noise-varying cross-domain samples. Furthermore, we argue that the conventional \( \mathcal{N}(\mathbf{0}, \mathbf{I}) \) initialization in diffusion models often generates class-confused hcpl-tds, compromising discriminative DA. To resolve this, we introduce a class-aware noise optimization strategy that refines sampling regions for reverse class-specific hcpl-tds generation, effectively enhancing cross-domain alignment. Extensive experiments across 5 benchmark datasets and 29 DA tasks demonstrate significant performance gains of \textbf{NOCDDA} over 31 state-of-the-art methods, validating its robustness and effectiveness.
\end{abstract}

\section{Introduction}
\label{Introduction}

The success of deep learning models relies heavily on abundant, well-labeled training data \((\mathcal{X}_\mathcal{S}, \mathcal{Y}_\mathcal{S})\), enabling impressive performance on test samples \((\mathcal{X}_\mathcal{T})\). These models typically assume that training and testing data are independent and identically distributed (\textit{iid}), implying the same statistical distribution for both domains. However, this assumption often breaks in real-world scenarios \cite{pan2010survey,DBLP:journals/csur/LuLHWC20}, where variations in environmental conditions (\textit{e.g.}, lighting or temperature) or data capture methods cause significant discrepancies between training and testing data. Such \textit{domain shifts} hinder the effectiveness of models trained on a specific labeled source domain (\(\mathcal{D}_{\cal S}\)) when applied to an unlabeled target domain (\(\mathcal{D}_{\cal T}\)). Unsupervised Domain Adaptation ({UDA}) addresses these challenges by leveraging models trained on \(\mathcal{D}_{\cal S}\) for direct application to \(\mathcal{D}_{\cal T}\), despite the inherent \textit{domain shift}.

The theoretical foundation of {UDA}, as established in Eq.\eqref{eq:1} \cite{ben2010theory,kifer2004detecting}, provides an error bound for the hypothesis \(h\) on the target domain, offering a framework to optimize models for cross-domain tasks under non-\textit{iid} conditions:
 \begin{equation}\label{eq:1}
 	\resizebox{0.8\hsize}{!}{%
 		$\begin{array}{l}
{e_\mathcal{T}}(h) \le \underbrace {{e_\mathcal{S}}(h)}_{{\bf{Term}}.{\bf{1}}} + \underbrace {{d_H}({\mathcal{D}_\mathcal{S}},{\mathcal{D}_\mathcal{T}})}_{{\bf{Term}}.{\bf{2}}} + \\
\underbrace {\min \left\{ {{E_{{\mathcal{D}_\mathcal{S}}}}\left[ {\left| {{f_\mathcal{S}}({\bf{x}}) - {f_\mathcal{T}}({\bf{x}})} \right|} \right],{E_{{\mathcal{D}_\mathcal{T}}}}\left[ {\left| {{f_\mathcal{S}}({\bf{x}}) - {f_\mathcal{T}}({\bf{x}})} \right|} \right]} \right\}}_{{\bf{Term}}.{\bf{3}}}
\end{array}$}
 \end{equation}

where the performance ${e_{\cal T}}(h)$ of a hypothesis $h$ on the \(\mathcal{D}_{\cal T}\) is bounded by three critical terms on the right-hand side. Specifically, \textbf{Term.1} represents the classification error on the \(\mathcal{D}_{\cal S}\). \textbf{Term.2} quantifies  the $\mathcal{H}$\emph{-divergence}\cite{kifer2004detecting} between the cross-domain distributions ($\mathcal{D_S}$, $\mathcal{D_T}$), and \textbf{Term.3} captures the discrepancy between the labeling functions of the two domains. In the context of UDA, optimizing {Term.1} is typically achieved through supervised learning to ensure effective functional regularization. However, optimizing {Term.2} and {Term.3} depends on hcpl-tds from \(\mathcal{D_T}\), which poses a challenge in UDA setups where labels from \(\mathcal{D_T}\) are unavailable.

\textbf{\textit{The scarcity of High-Confidence Pseudo-Labeled Target-Domain Samples (hcpl-tds).}}
Both \textbf{Term 2} and \textbf{Term 3} rely heavily on hcpl-tds for optimization. However, real-world domain shifts often restrict the availability of these samples, limiting the reduction of the error bound in Eq.\ (\ref{eq:1}).  


\textbf{Impact on Term.2}: The goal of Term.2 is to align cross-domain conditional distributions, \textit{i.e.}, \ 
\(
Q(\mathcal{Y}_S \mid \mathcal{X}_S)
\approx
Q(\mathcal{Y}_\mathcal{T} \mid \mathcal{X}_\mathcal{T}).
\)
In UDA, target domain pseudo-labels \(\mathcal{Y}_\mathcal{T}\) are often inaccurate due to domain shifts. To mitigate this, we select hcpl-tds to construct $\mathcal{D}_\mathcal{T}^h
=
\{(\mathcal{X}_\mathcal{T}^i,\,\mathcal{Y}_\mathcal{T}^i)\,\mid\,\mathrm{confidence}(\mathcal{Y}_\mathcal{T}^i)\,\ge\,\eta\},$
where \(\eta\) is a confidence threshold. However, \(\mathcal{D}_\mathcal{T}^h\) is typically much smaller than \(\mathcal{D}_\mathcal{T}\) (\textit{i.e.}, \(\left\|\mathcal{D}_\mathcal{T}^h\right\|_0 \ll \left\|\mathcal{D}_\mathcal{T}\right\|_0\)), limiting its ability to fully capture the statistical characteristics of \(\mathcal{D_T}\). This sparsity impedes accurate distribution alignment, ultimately affecting DA performance.

\textbf{Impact on Term 3.} Let \(f_\mathcal{S}(\mathbf{x})\) and \(f_\mathcal{T}(\mathbf{x})\) denote the source and target labeling functions, respectively. We define \(\mathcal{D}_\mathcal{T}^h = \{\mathbf{x} \in \mathcal{D}_\mathcal{T} \mid f_\mathcal{T}(\mathbf{x}) \ge \eta\},\) as the set of selected hcpl-tds, where \(f_\mathcal{T}(\mathbf{x})\) exceeds a predefined confidence threshold \(\eta\). Naturally, the size of \(\mathcal{D}_\mathcal{T}^h\) is often smaller than the total size of \(\mathcal{D}_\mathcal{T}\), limiting the effective sample space available for estimating the expectation in Term 3 of Eq.\eqref{eq:1}. Since \(\mathbb{E}_{\mathcal{D}_\mathcal{T}}\) depends on the target domain samples, this reduction compromises the quality of the estimated expectation \(\mathbb{E}_{\mathcal{D}_\mathcal{T}}\). Mathematically, let \(\hat{\mathbb{E}}_{\mathcal{D}_\mathcal{T}^h}[\|f_\mathcal{S}(\mathbf{x}) - f_\mathcal{T}(\mathbf{x})\|]\) denote the empirical expectation computed over \(\mathcal{D}_\mathcal{T}^h\). When \(\left\|\mathcal{D}_\mathcal{T}^h\right\|_0 \ll \left\|\mathcal{D}_\mathcal{T}\right\|_0\), the empirical estimation introduces a bias: \resizebox{0.75\hsize}{!}{$\bigg| \mathbb{E}_{\mathcal{D}_\mathcal{T}}[\|f_\mathcal{S}(\mathbf{x}) - f_\mathcal{T}(\mathbf{x})\|] - \hat{\mathbb{E}}_{\mathcal{D}_\mathcal{T}^h}[\|f_\mathcal{S}(\mathbf{x}) - f_\mathcal{T}(\mathbf{x})\|] \bigg| \to \varepsilon,$} where \(\varepsilon\) represents the error due to the insufficient representation of \(\mathcal{D}_\mathcal{T}\). This bias reduces the robustness of the alignment between \(f_\mathcal{S}(\mathbf{x})\) and \(f_\mathcal{T}(\mathbf{x})\), thereby hindering effective DA.

\begin{figure*}[h]
  \centering
\includegraphics[width=1\linewidth]{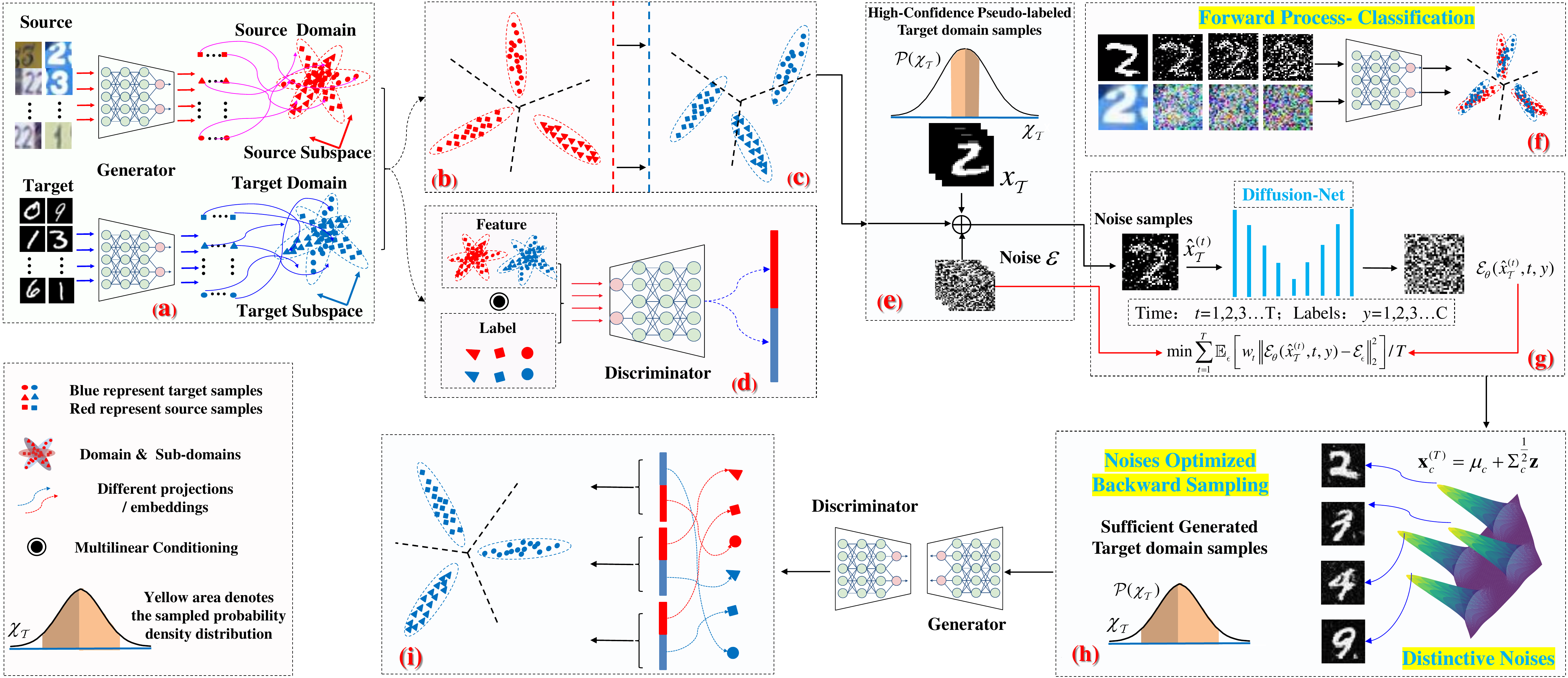}
\caption{Illustration of the proposed \textbf{NOCDDA} method. Fig.\ref{fig:1}(a) shows source (red) and target (blue) data with distinct distributions and geometric structures, where shapes indicate class labels. Fig.\ref{fig:1}(b\&c) depicts generator training on labeled source data and pseudo-label prediction for the target domain. Fig.\ref{fig:1}(d) aligns feature and label spaces via conditional adversarial learning. hcpl-tds selection is highlighted in Fig.\ref{fig:1}(e), followed by robust cross-domain classifier training under forward noise perturbation in Fig.\ref{fig:1}(f). Fig.\ref{fig:1}(g\&h) illustrates class-specific backward sampling and enriched target sample generation. Fig.\ref{fig:1}(i) shows enhanced DA performance via improved target representation.}
	\label{fig:1}
\end{figure*}

Recent studies have demonstrated the effectiveness of diffusion models \cite{song2020score,ho2020denoising,lipman2022flow} in sample generation and distribution inference. Building on this foundation, we propose a novel approach, \textbf{N}oise \textbf{O}ptimized \textbf{C}onditional \textbf{D}iffusion for \textbf{D}omain \textbf{A}daptation (\textbf{NOCDDA}), which seamlessly integrates diffusion-based generation with DA decision-making for synergistic optimization. Specifically, the forward diffusion process is coupled with DA classifier training, enabling robust cross-domain decision-making by injecting progressively increasing noise while refining cross-domain decision boundaries. Meanwhile, the reverse sampling phase employs noise-optimized strategies for class-aware \textbf{hcpl-tds} generation, enhancing both the quality and discriminative power of \(\mathcal{D}_\mathcal{T}\) representations for effective DA. \textit{In this framework, the superscript \((t)\) denotes the diffusion model’s time step, and the subscript \(\mathcal{T}\) represents the target domain.}

\textbf{\textit{Forward Diffusion Enabled Robust Cross-Domain Classifier Training}}: 
The forward diffusion process in {NOCDDA} incorporates two levels of constraints to achieve robust cross-domain classifier optimization.  
\textbf{Unified Optimization Framework:} By integrating the DA classifier with the conditional diffusion model’s classifier, a shared optimization framework is established, promoting invariant feature extraction and synergistic decision-making. \textbf{Noise-Enhanced Training:} The shared classifier is trained on sequentially noised samples from \(\mathcal{D}_{\cal S}\) and {hcpl-tds}, improving sensitivity to noise variations and adaptability in cross-domain scenarios. This dual-constraint strategy ensures that forward diffusion training not only reinforces cross-domain alignment but also enhances decision-making across diverse noise conditions, resulting in a robust and unified DA framework.

\textit{\textbf{Class-Specific Noise Optimization in Reverse Diffusion}}: 
In the reverse diffusion phase, traditional diffusion models initialize noise from \(P(\mathbf{x}^{(T)}) = \mathcal{N}(\mathbf{0}, \mathbf{I})\), assuming an infinite forward diffusion process (\(T \to \infty\)). However, real-world tasks with finite diffusion steps (\(T < \infty\)) yield terminal distributions \(P(\mathbf{x}^{(T)}) = \mathcal{N}(\boldsymbol{\mu}, \boldsymbol{\Sigma})\), where \(\boldsymbol{\mu}\) and \(\boldsymbol{\Sigma}\) represent the mean and covariance of noisy samples. To address this mismatch and meet the class-specific needs of domain adaptation (DA), we model class-specific terminal distributions tailored to hcpl-tds subsets. For each class \(c \in \mathcal{C}_\mathcal{T}\), the terminal distribution is defined as: \resizebox{0.88\hsize}{!}{$
P_c(\mathbf{x}^{(T)}) = \mathcal{N}(\boldsymbol{\mu}_c, \boldsymbol{\Sigma}_c), \boldsymbol{\mu}_c = \mathbb{E}[\mathbf{x}_c^{(T)}],  
\boldsymbol{\Sigma}_c = \mathrm{Cov}(\mathbf{x}_c^{(T)}),$}
where \(\boldsymbol{\mu}_c\) and \(\boldsymbol{\Sigma}_c\) are computed from hcpl-tds for class \(c\). This tailored initialization ensures reverse sampling trajectories align with class-aware hcpl-tds distributions, facilitating discriminative sample generation for robust cross-domain decision-making. Detailed visualizations in the supplementary material (\textcolor{blue}{Appendix.1}) validate the effectiveness of this approach across various diffusion models, illustrating how noise optimization significantly enhances class-specific sample generation. These findings highlight the critical role of tailored noise sampling in improving the efficiency and effectiveness of DA.

The key contributions are summarized as follows:

\begin{itemize}



\item \textbf{Forward Diffusion Enhanced Robust Classifier Training:} 
During the forward diffusion training process in the NOCDDA model, the DA classifier is modified and integrated with the diffusion model's classifier alongside a unified optimization framework. This coupling promotes robust cross-domain decision-making by enabling the DA classifier to distinguish cross-domain samples with varying noise intensities. Such integration effectively extracts invariant domain features, reinforcing domain alignment and further improving the model's robustness.

\item \textbf{Noise-Optimized Backward Sampling for Class-Specific hcpl-tds Generation:} 
Traditional diffusion models initialize reverse sampling with a uniform Gaussian prior \( \mathcal{N}(\mathbf{0}, \mathbf{I}) \), requiring an infinite forward diffusion process (\( T \to \infty \)), which is impractical for real-world tasks. This approach fails to meet the class-aware sample generation needs of DA. To address this, we propose a noise-optimized strategy, modeling class-specific terminal distributions \( P_c(\mathbf{x}^{(T)}) = \mathcal{N}(\boldsymbol{\mu}_c, \boldsymbol{\Sigma}_c) \), where \( \boldsymbol{\mu}_c \) and \( \boldsymbol{\Sigma}_c \) are computed from hcpl-tds under finite diffusion steps. This tailored initialization ensures reverse sampling aligns with hcpl-tds distributions, enabling class-aware generation and improving cross-domain decision-making in DA.

   \item We conducted extensive experiments on 15 image classification and 14 time series domain adaptation tasks across 31 comparison methods, showcasing the superior performance of our proposed {NOCDDA} method over state-of-the-art DA methods.

\end{itemize}

\section{Related Work}

Existing domain adaptation techniques are categorized into shallow and deep approaches. Given recent trends, this chapter focuses on deep DA methods to comparative analyze the rationale behind our proposed {NOCDDA} approach.

Traditional deep-DA techniques are generally classified into two paradigms: \textbf{non-adversarial} and \textbf{adversarial} methods. Non-adversarial methods rely on predefined statistical metrics such as Wasserstein distance \cite{tang2024source}, frequency spectrum \cite{yang2020fda}, and Maximum Mean Discrepancy (MMD) \cite{luo2024discriminative,long2018transferable} to quantify and minimize domain shift by reducing divergence in kernel feature spaces \cite{shu2024rul}. These methods depend on the careful selection and design of these metrics. In contrast, adversarial learning-based DA methods use adversarial training \cite{goodfellow2014generative} to align source and target domain features by making them indistinguishable through domain classifier training \cite{chen2023multicomponent,jing2024visually}. While this data-driven approach offers flexibility by removing the need for predefined metrics, it often results in unstable training due to the lack of explicit guidance for optimizing likelihood distributions \cite{bond2021deep,yang2022survey}. These DA techniques often focus on optimizing the mapping from samples to label space (\(f_d(\mathbf{x}, \theta) \mapsto p(\mathbf{c})\)), thus primarily operating as discriminative models.


In contrast to traditional discriminative models, which focus solely on inferring labels \(\mathbf{c}\), generative models (\(f_g(\theta) \mapsto p(\mathbf{x}, \mathbf{c})\)) have recently gained prominence in the DA field as an alternative approach. The key advantage of generative models lies in their ability to jointly model both the label distribution \(p(\mathbf{c})\) and the data distribution \(p(\mathbf{x})\) during the training process, preserving critical structural information within feature representations. This regularized functional learning framework \cite{yang2022survey,DBLP:conf/ijcai/JiangZTTZ17} enhances the robustness and reliability of generative approaches for DA. For example, Zhang \textit{et al.} \cite{zhang2020domain} framed DA as a Bayesian inference problem in probabilistic graphical models, while VDI \cite{xu2023domainindexing} extended this framework by incorporating domain indices into a variational Bayesian approach for multi-domain data. Recently, diffusion generative models \cite{song2020score,ho2020denoising,lipman2022flow}, leveraging Markov process-based sequential optimization, have emerged as a promising tool for DA, refining the joint probability distribution function \(p_\theta(\mathbf{x}_\mathcal{T}, \mathbf{x}_\mathcal{S})\) through reverse diffusion, facilitating smooth domain transitions and robust alignment.


Building on diffusion models, DAD \cite{peng2024unsupervised} bridges source and target domains by simulating a sequence of distributions with minimal discrepancies between adjacent steps, achieving seamless cross-domain blending. Trans-Diff \cite{kang2024trans} uses cross-domain prompts to generate semantically-aware pseudo-target domain images, significantly improving domain adaptation performance. DDA \cite{gao2023back} leverages a diffusion model to minimize domain divergence by aligning data structures in noisy latent spaces. Subsequently, Kamil \textit{et al.} \cite{deja2023learning} enhanced classifier guidance to improve conditional distribution-aware domain adaptation. Prof. Li \textit{et al.} \cite{du2024diffusion} further advanced cross-domain alignment through diffusion-based purification of target domain labels.

Existing diffusion models have advanced domain adaptation (DA) through sample augmentation and label optimization but often fail to integrate DA decision-making with generative objectives effectively. To address this limitation, we propose {NOCDDA}, a unified framework that seamlessly combines conditional diffusion generation with DA-specific objectives. In the forward diffusion phase, the joint classifier is trained on sequentially noised cross-domain data, enhancing its robustness and adaptability to domain discrepancies. In the reverse generation phase, {NOCDDA} incorporates noise-optimized sampling strategies, enabling class-specific hcpl-tds generation that aligns with cross-domain decision-making requirements. By unifying robust decision-making with class-aware sample augmentation, {NOCDDA} not only improves cross-domain alignment but also establishes a strong foundation for effective and discriminative DA performance.

\section{Noise Optimized Conditional Diffusion Motivated Domain Adaptation}

We begin by defining the notations used in this paper, followed by a clear explanation of the research motivation and a step-by-step breakdown of the proposed \textbf{NOCDDA} approach.

\subsection{Notations and Problem Statement}
\label{Notations and Problem Statement}

We use the cursive symbols $\mathcal{S}$ and $\mathcal{T}$ to represent the \emph{source} or \emph{target} domain, respectively. Matrices are denoted by boldface uppercase letters, \textit{e.g.}\ $\mathbf{M}\!=\!(m_{ij})$, whose $i$-th row is $\mathbf{m}^i$ and $j$-th column is $\mathbf{m}_j$. Vectors are denoted by boldface lowercase letters, \textit{e.g.} \ $\mathbf{x}$. A domain $\mathcal{D}$ is defined as an $\ell$-dimensional feature space $X$ together with a marginal distribution $P(\mathbf{x})$, \textit{i.e.}\ $\mathcal{D}=\{X,\,P(\mathbf{x})\}$ with $\mathbf{x}\in X$. Given a domain $\mathcal{D}$, a task $\mathcal{TK}$ consists of a $C$-cardinality label set $\mathcal{Y}$ and a classifier $f(\mathbf{x})$, \textit{i.e.}\ $\mathcal{TK}=\{\mathcal{Y},\,f(\mathbf{x})\}$, where $f(\mathbf{x})=\mathcal{Q}(y \mid \mathbf{x})$ interprets the class-conditional distribution over $\mathcal{Y}$. In unsupervised domain adaptation, we have a \emph{source domain} $\mathcal{D}_\mathcal{S} = \{\mathbf{x}_\mathcal{S}^i,\,\mathbf{y}_\mathcal{S}^i\}_{i=1}^{n_s}$ with $n_s$ labeled samples $\mathbf{X}_\mathcal{S}=[\,\mathbf{x}_\mathcal{S}^1\!\dots\!\mathbf{x}_\mathcal{S}^{n_s}]$ and labels $\mathbf{Y}_\mathcal{S}=\{\mathbf{y}_\mathcal{S}^1,\dots,\mathbf{y}_\mathcal{S}^{n_s}\}^\top \in \mathbb{R}^{n_s\times C}$, as well as an \emph{unlabeled target domain} $\mathcal{D}_\mathcal{T}=\{\mathbf{x}_\mathcal{T}^j\}_{j=1}^{n_t}$ with $n_t$ unlabeled samples $\mathbf{X}_\mathcal{T}=[\,\mathbf{x}_\mathcal{T}^1\!\dots\!\mathbf{x}_\mathcal{T}^{n_t}]$ and unknown labels $\mathbf{Y}_\mathcal{T}=\{\mathbf{y}_\mathcal{T}^1,\dots,\mathbf{y}_\mathcal{T}^{n_t}\}^\top \in \mathbb{R}^{n_t\times C}$. Although $X_\mathcal{S}=X_\mathcal{T}$ and $\mathcal{Y}_\mathcal{S}=\mathcal{Y}_\mathcal{T}$, the distributions and conditionals differ, \textit{i.e.}\ $P(X_\mathcal{S})\neq P(X_\mathcal{T})$, $\mathcal{Q}(\mathcal{Y}_\mathcal{S} \mid X_\mathcal{S})\neq \mathcal{Q}(\mathcal{Y}_t \mid X_\mathcal{T})$. We define a \emph{sub-domain} (class) in the source domain by $\mathcal{D}_\mathcal{S}^{(c)}=\{\mathbf{x}_\mathcal{S}^i\in\mathbf{X}_\mathcal{S} \mid \mathbf{y}_\mathcal{S}^i=c\}$. Sub-domains in the target domain, $\mathcal{D}_\mathcal{T}^{(c)}$, require a base classifier to assign pseudo labels to unlabeled samples in $\mathcal{D}_\mathcal{T}$. To distinguish the diffusion model's time step, we use the superscript $(t)$ for time encoding and the subscript $\mathcal{T}$ to denote the target domain.

\subsubsection{Time Embedding Aware Classifier Unification}
\label{Time Embedding Aware Classifier Unification}
As shown in Fig.\ref{fig:1}.b, to minimize \textbf{Term.1} in Eq.(\ref{eq:1}) for the source domain's Structural Risk Minimization, a well-trained classifier must be learned in a supervised manner. The hypothesis \(f: \mathcal{X}_\mathcal{S} \to \mathcal{Y}_\mathcal{S}\) maps source domain samples \(\mathcal{X}_\mathcal{S}\) to their labels \(\mathcal{Y}_\mathcal{S}\). However, such a hypothesis trained solely on this mapping fails to align with the classification decisions required in conditional diffusion training, which demands incorporating a time embedding \((t)\) to capture the temporal dynamics of the diffusion process. To address this, the predictive model is formulated as:
\begin{equation}\label{eq:srm}
\resizebox{0.9\hsize}{!}{$\begin{array}{l}
f = \arg \min_{f \in \mathcal{H}_{\mathcal{K}}} \mathbb{E}_{(x_\mathcal{S}, y_\mathcal{S}) \sim \mathcal{D}_\mathcal{S}} \Big[ l(f(x_\mathcal{S}, (t)), y_\mathcal{S}) \Big] + \mathcal{R}(f).
\end{array}$}
\end{equation}

Here, \(f(x_\mathcal{S}, (t))\) integrates the time embedding \((t)\), aligning the classifier's decisions with the conditional diffusion framework. For clean source domain images, \((t)\) is set to 0, ensuring consistency with the DA framework and maintaining alignment for noise-free data. This incorporation of time embedding within the DA classifier represents a key innovation, enabling unified decision-making across both DA and conditional diffusion training. The supervised loss \(l(f(x_\mathcal{S}, (t)), y_\mathcal{S})\) ensures accurate learning on the source domain, while the regularization term \(\mathcal{R}(f)\) controls the complexity of the hypothesis space \(\mathcal{H}_{\mathcal{K}}\). 


Despite the implemented  setup, cross-domain distribution divergence often leads to unreliable mappings for the target domain, \(f: (\mathbf{x}_{\mathcal{T}}, t) \mapsto \mathbf{y}_{\mathcal{T}}\). This underscores the need for effective strategies to minimize conditional distribution discrepancies and ensure efficient domain adaptation (DA).

\subsubsection{Adversarial Learning Enforced Domain Alignment}
\label{Adversarial Learning Enhanced Domain Alignment}
Despite Eq.(\ref{eq:srm}) effectively reducing the classification error on the \(\mathcal{D}_{\cal S}\) to minimize \textbf{Term.1} in Eq.(\ref{eq:1}), the learned hypothesis struggles to generalize to the \(\mathcal{D}_{\cal T}\) due to domain shift. To address this, we focus on optimizing \textbf{Term.2} by minimizing the cross-domain conditional distribution divergence. This is achieved by incorporating conditional adversarial training to align the hybrid conditional distributions of feature representations \(f\) and labels \(y\), as illustrated in Fig.\ref{fig:1}.(a-d). Specifically, Fig.\ref{fig:1}.(d) shows how cross-domain divergence is reduced through adversarial optimization of the \textbf{generator} (\(G\)) and \textbf{discriminator} (\({D}\)). The updated objective is formulated as:

\begin{equation}\label{eq:adv}
\resizebox{0.8\hsize}{!}{$\begin{array}{l}
\mathop {\min }\limits_G \mathop {\max }\limits_D {L_{adv}} = - \mathbb{E}[\sum\limits_{c = 1}^C {{\mathds{1}_{[{y_s} = c]}}\log \sigma (G({x_\mathcal{S}}, (t))} ] \\
+ \mathbb{E}[\log D({(f \otimes y)_\mathcal{S}})] + \mathbb{E}[\log (1 - D({(f \otimes y)_\mathcal{T}}))]
\end{array}$}
\end{equation}
In the RHS of Eq.(\ref{eq:adv}), the first term reformulates Eq.(\ref{eq:srm}), while the remaining two terms optimize \(G\) and \({D}\) within an adversarial framework to achieve model equilibrium. The symbol \(\otimes\) represents the multilinear conditioning operation \cite{long2018conditional}, which leverages the \(f\) and \(y\) inferred by \(G\) to perform kernel embedding for joint distribution alignment. This minimizes the conditional divergence between the source \((\mathcal{X}_\mathcal{S}, \mathcal{Y}_\mathcal{S})\) and target \((\mathcal{X}_\mathcal{T}, \mathcal{Y}_\mathcal{T})\) domains, facilitating smoother cross-domain knowledge transfer. However, conditional adversarial methods depend on High-Confidence Pseudo-labeled Target Domain Samples (hcpl-tds). In real-world settings, as shown in Fig.\ref{fig:1}.(c), target domain pseudo-labels are often noisy, leading to functional training errors. To address this, we quantify classifier prediction uncertainty using the entropy criterion:  \(E(\mathbf{f}) = -\sum_{c=1}^C f_c \log f_c\), where \(C\) is the number of classes, and \(f_c\) is the probability of assigning a sample to class \(c\). Samples with lower entropy are selected as hcpl-tds, thereby enhancing the reliability of training.

As discussed in Sec. (\ref{Introduction}), the limited availability of hcpl-tds fails to sufficiently represent the target domain's distribution or cover its discriminative decision space, hindering cross-domain alignment. To address this, we integrate conditional diffusion techniques to infer intrinsic distributions and generate diverse samples, thereby improving the target domain's representation and decision-making capacity, ultimately enabling robust and effective cross-domain adaptation.

\subsubsection{Forward Diffusion Enabled Robust Cross-Domain Classifier Training}
\label{Forward Noise Perturbation Encouraged Classifier Training}

Given the limited number of hcpl-tds, directly estimating and sampling from their probability density function is impractical. To overcome this, we employ conditional diffusion techniques with label guidance to infer the target domain's distribution. Specifically, our approach uses a forward diffusion process based on the DDPM framework \cite{ho2020denoising}, where noise is progressively added to the input data, training both the denoising model and the classifier. It is worth noting that the classifier \(f_\phi(\mathbf{x}, (t))\) is trained with \textbf{\textit{two levels of consistency constraints}}, optimizing the model specifically for DA tasks.

\noindent \textbf{\textit{1. Consistency Training With Noised Source and hcpl-tds Samples}}:
To facilitate target-domain sample generation through conditional diffusion, our classifier \(f_\phi\) learns from forward diffusion classification on hcpl-tds. To further enhance training, we also incorporate noised source-domain data, imposing \emph{consistency constraints} both \emph{across domains} (source vs.\ target) and \emph{across time} (multiple noise levels). Specifically, let \(\mathbf{x}^{(0)}\) be a clean sample (at time step \(0\)) drawn from the joint dataset \(\mathcal{D}_{\mathrm{joint}}\), which includes both source-domain data and hcpl-tds. We then minimize the following objective:

\begin{equation}\label{eq:f1}
\resizebox{0.9\hsize}{!}{
\(
   \min_\phi 
   \;\; \mathbb{E}_{\mathbf{x}^{(0)} \sim \mathcal{D}_{\mathrm{joint}}, \; t \sim \mathcal{U}(1, T)}
   \Bigl[
      \| f_\phi(\mathbf{x}_\mathcal{S}^{(t)}, (t)) - y_\mathcal{S} \|^2
      + \| f_\phi(\mathbf{x}_\mathcal{T}^{(t)}, (t)) - y_\mathcal{T} \|^2
   \Bigr],
\)
}
\end{equation}
where \(\mathbf{x}_\mathcal{S}^{(t)}\) and \(\mathbf{x}_\mathcal{T}^{(t)}\) represent sequentially noised versions of the source-domain sample \(\mathbf{x}_\mathcal{S}^{(0)}\) and hcpl-tds \(\mathbf{x}_\mathcal{T}^{(0)}\), respectively. By exposing the classifier \(f_\phi\) to varying noise intensities (\textit{i.e.}, different time steps \(t\)) for both domains, this training process enforces \emph{cross-time consistency} to improve robustness against noise, and \emph{cross-domain consistency} to facilitate domain-invariant feature learning. Consequently, the trained model achieves improved cross-domain generalization, providing a solid foundation for robust classifier training and adaptation to distribution shifts.

\noindent \textbf{\textit{2. Consistency Between the Diffusion-Based Classifier and the DA Classifier}}:
As discussed in Eq.(\ref{eq:srm}), the DA classifier is modified to share the same model parameters \(\phi\) as the diffusion-based classifier. Specifically, let \(f_\phi\colon \bigl(\mathbf{x}, t\bigr) \;\longmapsto\; \hat{y}\), where \(\hat{y}\) represents the predicted label (or label distribution) and \(t \in \{0, 1, \dots, T\}\) denotes the diffusion time step. The DA classifier corresponds to \(f_\phi(\mathbf{x}^{(0)}, 0)\), which operates on clean samples \(\mathbf{x}^{(0)}\) (\textit{i.e.}, \(t=0\)), while the diffusion-based classifier corresponds to \(f_\phi(\mathbf{x}^{(t)}, t)\) for \(t>0\), processing noised samples \(\mathbf{x}^{(t)}\) obtained from \(\mathbf{x}^{(0)}\) via the forward diffusion process. Since both classifiers use the same function \(f_\phi\), they naturally share label supervision. Specifically, let \(\{(\mathbf{x}_i^{(0)}, y_i)\}\) represent labeled samples from the joint dataset \(\mathcal{D}_{\mathrm{joint}}\). Our training objective combines losses over both clean (\(t=0\)) and noised (\(t>0\)) inputs:
\begin{equation}\label{eq:f2}
\resizebox{0.65\hsize}{!}{$\begin{array}{l}
\min_{\phi} 
\;\; 
\underbrace{
\mathbb{E}_{\mathbf{x}^{(0)} \sim \mathcal{D}_{\mathrm{joint}}}
\bigl[\| f_\phi(\mathbf{x}^{(0)}, 0) - y \|^2 \bigr]
}_{\text{DA on clean samples}}
\;+ \\ \;
\underbrace{
\mathbb{E}_{\mathbf{x}^{(0)} \sim \mathcal{D}_{\mathrm{joint}}, \; t \sim \mathcal{U}(1, T)}
\bigl[\| f_\phi(\mathbf{x}^{(t)}, t) - y \|^2 \bigr]
}_{\text{Diffusion on noised samples}}.
\end{array}$}
\end{equation}

Here, \(\mathbf{x}^{(t)}\) is the noisy version of \(\mathbf{x}^{(0)}\), and \(y\) is the same label supervision applied in both the DA (clean) and diffusion (noised) scenarios. While we do not explicitly impose a constraint like  
\(\| f_\phi(\mathbf{x}^{(t)}, t) - f_\phi(\mathbf{x}^{(0)}, 0) \|^2\), the shared parameterization \(\phi\) inherently promotes \emph{temporal continuity} (across different \(t\)-values) and \emph{mutual reinforcement} between the two tasks. Thus, improvements achieved on noised inputs (\(t>0\)) transfer to the clean case (\(t=0\)), and vice versa, enabling robust cross-domain alignment and enhancing overall adaptation performance.

By jointly optimizing Eq.(\ref{eq:f1}) and Eq.(\ref{eq:f2}), the proposed forward diffusion training ensures robust cross-domain decision alignment and temporal consistency between the DA and diffusion classifiers, thereby laying a solid foundation for effective reverse conditional sampling in hcpl-tds generation and enhanced adaptation.

\subsubsection{Noise Optimized Backward Generation}
\label{Efficient Backward sampling Augmented Data Distribution}

Building on the trained denoising model and the jointly optimized classifier, we adopt a reverse conditional generation strategy inspired by BeatGAN \cite{dhariwal2021diffusion} to enhance hcpl-tds quantity and leverage DDIM \cite{song2020denoising} to reduce reverse diffusion steps. Traditional diffusion methods sample from a Gaussian prior \( \mathcal{N}(\mathbf{0}, \mathbf{I}) \), assuming an infinite forward diffusion process (\( T \to \infty \)), which is impractical for real-world tasks. Finite diffusion steps (\( T < \infty \)) lead to deviations in the terminal distribution, causing class overlap in generated hcpl-tds and hindering effective cross-domain decision-making (\textit{see \textcolor{blue}{Appendix.1}}). To address this, we model class-specific terminal distributions \( \mathcal{N}(\boldsymbol{\mu}_c, \boldsymbol{\Sigma}_c) \), where \( \boldsymbol{\mu}_c \) and \( \boldsymbol{\Sigma}_c \) are derived from hcpl-tds under finite forward diffusion. This noise-optimized approach ensures class-aware, discriminative hcpl-tds generation, improving \( \mathcal{D}_{\mathcal{T}} \) representation and cross-domain alignment for domain adaptation tasks.

\textbf{\textit{1.Class-Specific Terminal Distributions:}}
For \(C\) classes in \(\mathcal{D}_{\cal T}\), we partition the hcpl-tds into subsets \(\{\mathcal{D}_\mathcal{T}^{(1)}, \dots, \mathcal{D}_\mathcal{T}^{(C)}\}\), where each subset corresponds to samples pseudo-labeled as class \(c\). Instead of assuming a shared Gaussian distribution for all classes at \(t = T\), we estimate class-specific terminal distributions as:  $\mu_c = \mathbb{E}[\mathbf{x}_c^{(T)}], \quad \Sigma_c = \mathrm{Var}[\mathbf{x}_c^{(T)}],$ where \(\mathbf{x}_c^{(T)} \in \mathcal{D}_\mathcal{T}^{(c)}\) represents the noised sample at the final forward step \(T\). However, due to uniform noise injection during forward diffusion, \(\Sigma_c\) tends to be large and similar across classes, leading to overlapping distributions. To mitigate this, we apply a regularization strategy that uniformly scales down the variances as:  $\Sigma_c = \frac{1}{C}\,\mathbf{I},$ where \(C\) is the number of classes. This adjustment minimizes inter-class overlap, enabling precise noise optimization and improving the quality of hcpl-tds generation.

\textbf{\textit{2. Noise Optimized Reverse Sampling:}}  For class-specific generation, we incorporate class-specific priors \((\mu_c, \Sigma_c)\) into the reverse sampling process, refining both the initialization and backward updates within the DDIM framework. Mathematically, let \(\epsilon_\theta\) denote the trained noise-prediction network, and \(p_\phi(y \mid \mathbf{x}^{(t)})\) represent the classifier’s label distribution function.

\textbf{Case 1}: Initialization at \(t = T\) (Class-Specific Starting Point)  
The terminal state of the forward diffusion process at \(t = T\) is modeled as a class-specific Gaussian distribution:
\begin{equation}\label{eq:b1}
\resizebox{0.8\hsize}{!}{$
\mathbf{x}_c^{(T)} 
\sim 
\mathcal{N}(\mu_c, \Sigma_c),
\quad s.t.~ 
\mu_c = \mathbb{E}[\mathbf{x}_c^{(T)}], 
~ \Sigma_c = \tfrac{1}{C}\,\mathbf{I}.
$}
\end{equation}
Here, \(\mu_c\) represents the mean of noised samples for class \(c\), while the scaled variance \(\Sigma_c = \tfrac{1}{C}\,\mathbf{I}\) reduces inter-class overlap by decreasing variance as the number of classes \(C\) increases. This adjustment ensures distinct reverse sampling trajectories, avoiding the overlap that would occur with a uniform variance (\(\Sigma_c = \mathbf{I}\)). In practice, the initialization is performed as:
\begin{equation}\label{eq:b2}
\resizebox{0.75\hsize}{!}{$\mathbf{x}_c^{(T)} 
= 
\mu_c 
+ 
\Sigma_c^{\frac{1}{2}} \mathbf{z},\quad 
s.t.~ 
\mathbf{z} \sim \mathcal{N}(\mathbf{0}, \mathbf{I}), ~ \Sigma_c = \tfrac{1}{C}\,\mathbf{I}.$}
\end{equation}
This initialization anchors the reverse sampling process to the empirical distribution of hcpl-tds for each class \(c\), ensuring the reverse trajectory begins within a class-specific region, thereby enabling class-aware hcpl-tds generation.

\textbf{Case 2}: Reverse Sampling for \(t < T\) (Adapted from DDIM with Class-Specific Initialization).  
Unlike standard DDIM, which initializes reverse diffusion with \(\mathbf{x}^{(T)} \sim \mathcal{N}(\mathbf{0}, \mathbf{I})\), we begin from a class-specific distribution \( \mathbf{x}_c^{(T)} \sim \mathcal{N}(\mu_c, \Sigma_c),\) as defined in Eq.\,\eqref{eq:b1}--\eqref{eq:b2}. This initialization ensures that the entire reverse trajectory remains \emph{class-aware}. At each step \(t \to t-1\), the DDIM-style update is enhanced by incorporating the classifier gradient to align with target-domain decisions. Specifically, the \emph{classifier-guided} noise term is defined as:
\begin{equation}\label{eq:b3}
\resizebox{0.9\hsize}{!}{$
\hat{\epsilon}_c(\mathbf{x}^{(t)}) 
= 
\epsilon_\theta(\mathbf{x}^{(t)}, t) 
- 
\sqrt{1 - \overline{\alpha}_t}\;\nabla_{\mathbf{x}^{(t)}} \log p_\phi(y \mid \mathbf{x}^{(t)}),
$}
\end{equation}
where \(\epsilon_\theta(\mathbf{x}^{(t)}, t)\) predicts the noise component, and the classifier gradient term \(-\nabla_{\mathbf{x}^{(t)}} \log p_\phi(y \mid \mathbf{x}^{(t)})\) steers the samples toward more discriminative regions.

\textit{Backward Update.}  
Starting from \(\mathbf{x}_c^{(T)}\) at \(t = T\), the state at each subsequent step \(t > 0\) is updated as:
\begin{equation}\label{eq:b4}
\resizebox{0.88\hsize}{!}{$
\mathbf{x}_c^{(t-1)} 
= 
\sqrt{\overline{\alpha}_{t-1}} 
\Bigl(\,
   \tfrac{\mathbf{x}_c^{(t)} - \sqrt{1 - \overline{\alpha}_t}\,\hat{\epsilon}_c(\mathbf{x}_c^{(t)})}{\sqrt{\overline{\alpha}_t}}
\Bigr)
+ 
\sqrt{1 - \overline{\alpha}_{t-1}}\,\hat{\epsilon}_c(\mathbf{x}_c^{(t)}),
$}
\end{equation}
where each \(\mathbf{x}_c^{(t)}\) recursively depends on the previous state \(\mathbf{x}_c^{(t+1)}\). Overall, this class-specific initialization and classifier-guided reverse sampling strategy ensure that the process aligns with the empirical terminal distribution of each class. This approach significantly enhances the generation of class-specific hcpl-tds, facilitating improved cross-domain separability and discriminative adaptation.


\textbf{\textit{3. Merits and Discussion of Noise-Optimized Conditional Generation:}}  
As detailed in the supplementary material (\textcolor{blue}{Appendix.1}), we evaluate three diffusion-based approaches, namely DSM \cite{vincent2011connection}, Consistency Distillation, and Consistency Training \cite{song2023consistency}, to validate the effectiveness of noise optimization for class-aware sample generation. Visualized reverse sampling trajectories demonstrate that incorporating class-specific terminal distributions \(\mathcal{N}(\mu_c, \tfrac{1}{C}\,\mathbf{I})\) under finite forward diffusion steps facilitates decision-boundary-aware hcpl-tds generation. Although constraining terminal noise may slightly limit generative diversity, it aligns with the fundamental objective of DA: optimizing cross-domain decision-making. To sum up, by resolving the inherent conflict between finite diffusion steps and unified noise distributions, the proposed noise-optimized strategy prioritizes classification-driven objectives, thereby enhancing cross-domain alignment and improving DA performance.


\subsubsection{NOCDDA Motivated Domain Adaptation:}
\label{NOCDDA Motivated Domain Adaptation:}
In summary, the proposed \textbf{NOCDDA} strategy is specifically tailored for DA, integrating robust cross-domain classifier optimization in the forward diffusion process with class-aware hcpl-tds generation in the reverse sampling process. By coupling conditional diffusion with DA models, NOCDDA achieves synergistic optimization, effectively enhancing cross-domain decision boundary alignment.


\section{Experiments}
The experimental section covers Dataset Description, Experimental Setup, 31 Baseline Methods, Experimental Results and Discussion, and includes an Ablation Study to further discuss the individual contributions of the proposed method's design.

	\subsection{ Dateset Description}
	\label{subsection:Benchmarks and Features}

\textbf{Image datasets}: \textbf{\textit{Digits}}: We evaluate domain adaptation on three digit datasets: {MNIST} ({M}) with 60,000 training and 10,000 test samples, {USPS} ({U}) with 7,291 training and 2,007 test samples, and {SVHN} ({S}) with over 600,000 labeled street view digits.  \textbf{\textit{Office-31}}: This benchmark dataset includes over 4,000 images across three domains: \textit{Amazon} ({A}), \textit{Webcam} ({W}), and \textit{Dslr} ({D}), used to evaluate six transfer tasks such as \textit{A} $\rightarrow$ \textit{W} and \textit{W} $\rightarrow$ \textit{D}.  \textbf{\textit{ImageCLEF-DA}}: This dataset contains 12 shared classes from \textit{Caltech-256} ({C}), \textit{ImageNet ILSVRC 2012} ({I}), and \textit{Pascal VOC 2012} ({P}), covering six cross-domain tasks like \textit{C} $\rightarrow$ \textit{I} and \textit{I} $\rightarrow$ \textit{P}.  \textbf{Time-series datasets}: \textit{\textbf{CWRU}}: The Case Western Reserve University (CWRU) dataset includes vibration data across four operating speeds (domains A, B, C, D) and 10 health states, enabling 12 cross-domain tasks.  \textit{\textbf{SEU}}: The Southeast University (SEU) dataset includes bearing and gear data under two working conditions (tasks A and B) with imbalanced fault sampling, providing a challenging DA benchmark. \textit{Detailed dataset descriptions are available in \textcolor{blue}{Appendix.2} of the supplementary material.}

	\subsection{Experimental Setup}
	\label{subsection: Experimental setup}

Experiments were conducted on several datasets, including Digits, Office-31, ImageCLEF-DA, CWRU, and SEU, using the PyTorch framework and an Nvidia 4090 GPU. For the Digits dataset, the diffusion model was trained with 1000 diffusion steps using DDIM sampling (200-step schedule with 5-step jumps), a batch size of 36, a learning rate of 0.02, and momentum of 0.5 for 100 epochs. For the Office-31 and ImageCLEF-DA datasets, DDIM sampling was optimized with 50 steps (20-step jumps), generating 100 images per class, with learning rates of 0.001 (for Office-31) or 0.0003 (depending on the domain pair). For CWRU and SEU, DDIM with 10 steps (100-step jumps) was used, generating 50 samples per class, with a batch size of 64, a learning rate of 0.03, and 50 epochs for CWRU and 30 for SEU. The model employed a U-Net architecture (1D for CWRU and SEU, 2D for Office-31 and ImageCLEF-DA), incorporating encoder-decoder structures, skip connections, attention layers, and residual blocks. \textit{For detailed settings, please refer to \textcolor{blue}{Appendix.3} in the supplementary material.} Model performance was evaluated on the target domain test set \({\cal{D_T}}\) using accuracy, as defined in Eq.\ (\ref{eq:accuracy}), a widely-used metric in prior research \cite{long2018conditional,li2024principal}.

\begin{equation}\label{eq:accuracy}
	Accuracy = \frac{{\left| {x \in {\cal{D_T}} \wedge \hat y(x) = y(x)} \right|}}{{\left| {x \in {\cal{D_T}}} \right|}},
\end{equation}

where \(\hat{y}(x)\) is the predicted label, and \(y(x)\) is the ground truth label for test data \(x\).

	\subsection{Baseline Methods}
	\label{subsection:Baseline Methods}
The developed {NOCDDA} method was evaluated against \textbf{31} established DA techniques, including  
\textbf{DOLL} \cite{luo2024discriminative}, 
\textbf{TLSR} \cite{luo2024semi}, 
\textbf{MCADA} \cite{chen2023multicomponent}, 
\textbf{CaCo} \cite{huang2022category}, 
\textbf{SUDA} \cite{zhang2022spectral}, 
\textbf{SATLN} \cite{qin2022deep}, 
\textbf{PfAReLU} \cite{chen2023domain}, 
\textbf{ATM} \cite{li2020maximum},
\textbf{TPN} \cite{pan2019transferrable}, 
\textbf{BSP} \cite{chen2019transferability}, 
\textbf{CAT} \cite{deng2019cluster}, 
\textbf{MSTN} \cite{xie2018learning}, 
\textbf{MCD} \cite{saito2018maximum}, 
\textbf{MADA} \cite{pei2018multi},
\textbf{GTA} \cite{sankaranarayanan2018generate}, 
\textbf{CDAN} \cite{long2018conditional}, 
\textbf{CyCADA} \cite{pmlr-v80-hoffman18a}, 
\textbf{UNIT}, 
\textbf{ADDA}, 
\textbf{JAN},
\textbf{DRCN}, 
\textbf{CoGAN}, 
\textbf{DANN},
\textbf{DAN}, 
\textbf{DDC},
\textbf{D-CORAL},
\textbf{RTN},
\textbf{MK-MMD},
\textbf{JMMD},
as well as baseline models such as \textbf{ResNet} and \textbf{CNN}. \textit{For full references of the compared models, please refer to \textcolor{blue}{Appendix.4} of the supplementary material.} Performance metrics for these methods were obtained from their original publications or subsequent analyses \cite{luo2024discriminative,chen2023domain}, reflecting their reported best results.

	\subsection{Experimental Results and Discussion}
	\label{subsection: Experimental Results and Discussion}	
\begin{figure}[h!]
		\centering
		\includegraphics[width=1\linewidth]{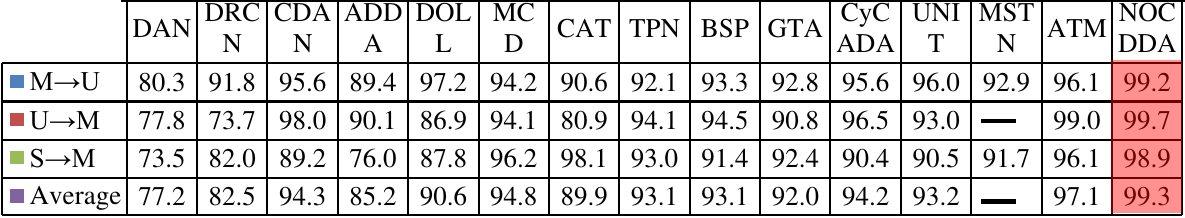}
		\caption {Accuracy${\rm{\% }}$ on Digital Image Datasets.} 
		\label{fig:Digits}
	\end{figure}

\begin{figure}[h!]
		\centering
		\includegraphics[width=1\linewidth]{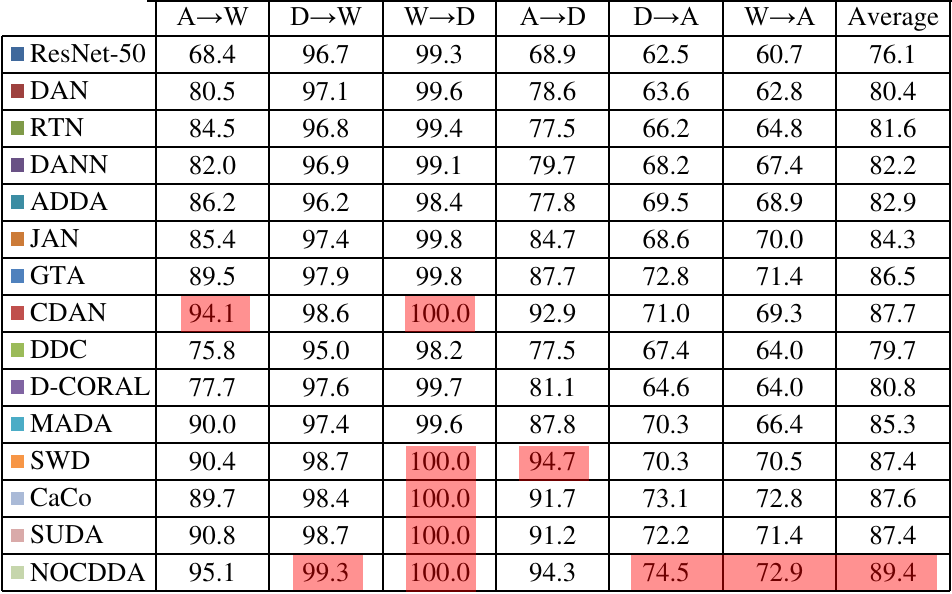}
		\caption {Accuracy${\rm{\% }}$ on Office-31 Datasets.} 
		\label{fig:office31}
	\end{figure}

\textbf{Digits}: Fig.\ref{fig:Digits} shows the DA results on the digit datasets, with the best results highlighted in red. Among existing methods, {ATM} \cite{li2020maximum}, which builds on {CDAN}, achieves a second-best accuracy of \(97.1\%\) by improving domain separation. In comparison, {NOCDDA} achieves the highest accuracy of \(99.3\%\) by coupling the forward diffusion process with DA decision-making, ensuring robust cross-domain consistency. The reverse generation process further enhances the target domain’s statistical representation through optimized noise sampling. \textbf{Office-31}: Fig.\ref{fig:office31} compares popular DA methods on the {Office-31} dataset, which includes 31 office-related categories. Using {ResNet} as a baseline, methods like {ADDA} align features through adversarial learning, while {MADA} refines decision boundaries with multiple discriminators. In contrast, {NOCDDA} optimizes cross-domain representations through dual-classifier consistency and noise-optimized reverse generation, achieving the highest average accuracy of \(89.4\%\). \textit{Experimental results and detailed discussions for \textbf{ImageCLEF-DA} and cross-domain time-series datasets \textbf{SEU} and \textbf{CWRU} are provided in \textcolor{blue}{Appendix.5, 6 \& 7}, respectively.}

\vspace{-5mm}

	\subsection{\textbf{Ablation Study}}
	\label{Empirical Analysis}

\begin{figure}[h!]
		\centering
		\includegraphics[width=0.9\linewidth]{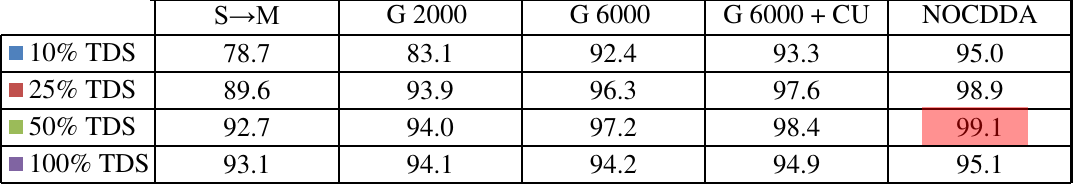}
		\caption {Ablation Study on Digital Image Datasets.} 
		\label{fig:ablation}
	\end{figure}

Although the {NOCDDA} method achieves state-of-the-art performance across 29 DA tasks in 5 datasets, outperforming 31 existing methods, its success hinges on three key components: \textit{hcpl-tds selection}, \textit{classifier unification-enhanced forward training}, and \textit{noise-optimized reverse generation}. To evaluate the contributions of each component, an ablation study was conducted on the \textit{SVHN-to-MNIST} DA task. As shown in Fig.~\ref{fig:ablation}, the vertical axis, ‘{X\% TDS}’, represents the proportion of high-confidence, low-entropy target domain samples, while the horizontal axis indicates the number of generated samples (\textit{e.g.}, \(G6000\) corresponds to 6000 samples generated using the diffusion model). The \(G6000+CU\) model integrates classifier unification in the forward diffusion phase, while {NOCDDA} includes noise optimization during reverse sampling. Key observations include: \textbf{1. Full TDS Utilization (\(TDS = 100\%\))}: Utilizing all target domain samples achieves a baseline accuracy of \(93.1\%\), demonstrating the importance of TDS for effective \(\mathcal{D}_\mathcal{T}\) modeling. However, doubling the sample size from \(TDS = 50\%\) to \(TDS = 100\%\) yields only a marginal 0.4\% improvement, revealing the limited contribution of low-confidence samples. This underscores the critical role of hcpl-tds in enhancing discriminative DA by prioritizing high-confidence target samples for cross-domain alignment. \textbf{2. Impact of Generated Samples}: At \(TDS = 10\%\), generating 6000 additional samples (\(G_{6000}\)) boosts baseline accuracy by \(+13.7\%\) (from \(78.7\%\) to \(92.4\%\)), demonstrating the importance of  hcpl-tds generation in enhancing data representation for discriminative DA. \textbf{3. Joint Optimization (\(G_{6000} + CU\))}: Incorporating classifier unification (\(CU\)) into the forward diffusion training further improves performance. For example, at \(TDS = 50\%\), accuracy increases from \(97.2\%\) (\(G_{6000}\)) to \(98.4\%\) (\(G_{6000} + CU\)), validating the efficacy of coupling DA classification with the generative model’s conditional diffusion. \textbf{4. Noise Optimization}: \textbf{NOCDDA} achieves the highest accuracy of \(99.1\%\) at \(TDS = 50\%\), demonstrating that noise-optimized sampling (\(\mathcal{N}(\mu_c, \Sigma_c)\)) effectively enhances class-specific hcpl-tds generation. This further refines cross-domain decision-making and underscores the importance of noise optimization in DA tasks.
\vspace{-5mm}
    
\vspace{-5mm}

	\section{Conclusion}
	\label{Conclusion}

In this research, we propose Noise Optimized Conditional Diffusion for Domain Adaptation (NOCDDA), a novel approach to enhancing cross-domain functional learning. Unlike traditional diffusion-based DA methods, NOCDDA seamlessly integrates DA decision-making with the generative process of conditional diffusion, fostering mutual reinforcement between the two tasks. During the forward diffusion phase, the classifier is unified with the DA decision-maker, enabling robust cross-domain decision training. In the reverse sampling phase, noise optimization ensures the generation of class-specific hcpl-tds, enhancing discriminative performance. Experimental results on five benchmark datasets, along with comparisons against 31 state-of-the-art methods, validate the effectiveness and competitiveness of NOCDDA in DA tasks.

\bibliographystyle{named}
\bibliography{ijcai25}

\end{document}